%% file: main.tex
\pdfoutput=1
\documentclass{article} %
\usepackage{iclr2023_conference,times}
\usepackage{acronym}

\usepackage{url}
\usepackage{booktabs}
\usepackage{caption}
\usepackage{float}
\usepackage{titlesec}
\usepackage{capt-of}
\usepackage{comment}
\usepackage{adjustbox}
\usepackage{xspace}

\usepackage{graphicx}
\usepackage{subfig}
\input{math_commands.tex}

\input{authors.tex}

\title{Evaluating the Robustness of Machine\\ Reading Comprehension Models to Low\\ Resource Entity Renaming}

\newcommand{\EntSwap}{Entity Swapping}

\iclrfinalcopy %
\begin{document}

\maketitle
\begin{abstract}

Question answering (QA) models have shown compelling results in the task of Machine Reading Comprehension (MRC). Recently these systems have proved to perform better than humans on held-out test sets of datasets e.g. SQuAD, but their robustness is not guaranteed. The QA model's brittleness is exposed when evaluated on adversarial generated examples by a performance drop. In this study, we explore the robustness of MRC models to entity renaming, with entities from low resource regions such as Africa. We propose EntSwap, a method for test-time perturbations, to create a test set whose entities have been renamed. In particular, we rename entities of type: \textit{country, person, nationality, location, organization} and \textit{city}, to create AfriSQuAD2. Using the perturbed test set, we evaluate the robustness of three popular MRC models. We find that compared to base models, large models perform well comparatively on novel entities. Furthermore, our analysis indicate that \textit{person}, as an entity type, highly challenges the model performance.
\end{abstract}

\input{sections/01-introduction.tex}

\input{sections/02-relatedwork.tex}

\input{sections/03-Method.tex}

\input{sections/04-entswap.tex}
\input{sections/05-results.tex}

\input{sections/06-conclusion.tex}

\subsubsection*{Acknowledgments}
We thank our reviewers for the valuable feedback. This research was partly supported by the Dreams Lab, a collaboration between Huawei, the University of Amsterdam, and the Vrije Universiteit Amsterdam, and the Science Foundation Ireland (SFI)\footnote{\url{https://www.sfi.ie/}} under Grant Number SFI/12/RC/2289\_P2, co-funded by the European Regional Development Fund.

All content represents the opinion of the authors, which are not necessarily shared or endorsed by their respective employers and/or sponsors.

\bibliography{references}
\bibliographystyle{iclr2023_conference}

\end{document}

%% file: math_commands.tex
\usepackage{amsmath,amsfonts,bm}

\def\eqref#1{equation~\ref{#1}}

\def\1{\bm{1}}

\DeclareMathAlphabet{\mathsfit}{\encodingdefault}{\sfdefault}{m}{sl}
\SetMathAlphabet{\mathsfit}{bold}{\encodingdefault}{\sfdefault}{bx}{n}

%% file: authors.tex
\author{Clemencia Siro \\
University of Amsterdam \\
Amsterdam, The Netherlands \\
\texttt{c.n.siro@uva.nl} \\
\AND
Tunde Oluwaseyi Ajayi \\
Insight Centre for Data Analytics \\
University of Galway \\
Galway, Ireland \\
\texttt{tunde.ajayi@insight-centre.org}
}

%% file: sections/01-introduction.tex
\section{Introduction}

Machine Reading Comprehension (MRC) is a question-answering task over unstructured text with the aim of examining the understanding and reasoning capability of a model. Over the past few years, there has been growing interest in this task due to the availability of large-scale datasets such as SQuAD \citep{rajpurkar-etal-2016-squad}, MS MARCO~\citep{DBLP:conf/nips/NguyenRSGTMD16}. Furthermore, the advent of deep learning techniques and frameworks~\citep{ NIPS2015_8fb21ee7, DBLP:conf/nips/VaswaniSPUJGKP17, devlin-etal-2019-bert} has improved the performance of MRC models as shown in some model performance on some specific tasks as compared to humans~\citep{rajpurkar-etal-2016-squad}. 

Despite the impressive performance, these models show poor performance on adversarial attacks compared to humans. Given the results in recent works, the MRC models are still not robust to adversarial attacks on all natural language understanding (NLU) tasks~\citep{jia-liang-2017-adversarial, DBLP:conf/iclr/BelinkovB18} and out-of-distribution examples~\citep{DBLP:conf/acl/TalmorB19,DBLP:conf/blackboxnlp/McCoyML20}.
 Several works have proposed investigating the robustness of MRC models to test-time perturbations by creating adversarial examples~\citep{jia-liang-2017-adversarial}. One of the earlier works by ~\citep{jia-liang-2017-adversarial}  appended semantically irrelevant sentences containing a fake answer that resembles the question syntactically to the context to confuse the model. The authors show how fragile SQuAD models are with the introduction of out-of-distribution phenomena whereby perturbing the test set yields close to a 50\% drop in generalization performance. 
 
 The notion of MRC robustness has been investigated in several different settings~\citep{DBLP:conf/aaai/JinJZS20, DBLP:conf/acl/SiYCMLW21,DBLP:conf/naacl/LiuL0S22}. Recently~\citet{DBLP:conf/naacl/YanXMLJR22} created adversarial examples by renaming entity names in several datasets with novel entities. The authors proved there is a discrepancy in model performance between entities in the answers observed during training and novel answers. Our work builds on this idea to investigate the robustness of MRC models in renaming English entities with African-based entities. We leverage the method of entity swapping to create a test dataset\footnote{https://github.com/}. In particular, we investigate the distribution shift at test-time caused by entities (e.g., country and city) with names from African region.  Using existing MRC datasets, SQuAD2.0~\citep{rajpurkar-etal-2018-know} we create adversarial examples to evaluate the robustness of span-based MRC models to test-time perturbations.

 A robust model, even though it has observed a small subset of all possible entity names available, should be able to generalize to novel entities.
Though simple and understudied, entity swapping tests the model's capability to generalize to novel entities due to a large number of possible entities. In addition, an entity name has world knowledge associated with it and this may change at any given time. Thus, MRC models should not overly rely on specific entities as this would lead to poor generalization on novel entities. Therefore in this study, we first investigate the distribution of entity names from both the train and dev set of SQuAD2.0. We show that the most common entity names are from high-resourced regions~(e.g., Europe, America, etc.) compared to Africa. As such, we investigate the robustness of MRC models in answering questions and extracting answers with entity names from Africa. 

Our key contributions are as follows: 1) We propose a method to create adversarial examples with entities from low-resource regions such as Africa. Since most of these regions have fewer digitized articles on Wikipedia, this method can be used to ensure a fair representation of regions during dataset creation. 2) We provide a detailed analysis on the robustness of MRC models to entity names from Africa. We show that although large models generalize comparatively well to novel entities, there is still a performance drop. 3) In our error analysis, we highlight the factors affecting the model's performance, thereby limiting its robustness to entity renaming, which we believe will foster future research towards more robust MRC models.

%% file: sections/02-relatedwork.tex
\section{Related Works}

The use of adversarial examples to evaluate and improve the robustness of machine learning models has a long-standing history \citep{holmstrom1992using, DBLP:conf/nips/WagerWL13}. In the field of NLP, one of the early works by~\citet{jia-liang-2017-adversarial}  showed that despite existing neural network QA systems proving their success when evaluated on standard metrics, they perform poorly when evaluated on adversarial examples. In their work, they propose the creation of adversarial examples for SQuAD v1.1 using the AddSent and AddAny algorithms. In AddSent, a distractor sentence is appended at the end of each context. For the AddAny algorithm, a random sequence of grammatical and ungrammatical words are appended to each context. They retrained the BIDAF model \citep{DBLP:conf/iclr/SeoKFH17} on these generated adversarial examples to test its robustness. AddSent algorithm swaps named-entities and numbers in the question with the nearest word in GloVe word vector space \citep{manning-EtAl:2014:P14-5}. Our method, EntSwap, slightly differs from AddSent. We replace named-entities, while leaving numbers unchanged. Unlike AddSent, which replaces entities in the question, appends distracting sentences to the context, and leaves the answers unchanged, EntSwap replaces all detected named-entities for the questions, context, and answers to create an altered SQuAD2.0 dev set.

Although the BIDAF model \citep{DBLP:conf/iclr/SeoKFH17} was retrained on adversarial examples, there is no guarantee of its robustness when evaluated on adversarial examples generated differently. Wang and Bansal~\citep{DBLP:conf/naacl/WangB18} generated slightly different adversarial examples for SQuAD using the AddSentMod algorithm by prepending the distractor sentences to the context instead of appending them and also used a different set of fake answers from ADD-SENT. The authors show that the pre-trained BiDAF model \citep{DBLP:conf/iclr/SeoKFH17} is not robust to this set of adversarial examples as the model’s F1-score drops by $30\%$. While both works use distractor sentences to create adversarial examples, our study randomly swaps English named-entities with entity names of African origin.

Similar to our work, \citet{DBLP:conf/naacl/YanXMLJR22}, creates adversarial examples by renaming entity names in several datasets, including SQuAD, with novel entities. The authors' perturbation involved detecting an entity in the answer span and then swapping all the occurrences of that entity in the passage for the categories: person names, organizations, and geopolitical entities. Our work differs from~\citep{DBLP:conf/naacl/YanXMLJR22} in the following ways: we swap six categories of entity types: person, city, country, organization, nationality, and location. Every entity swapped in the context is swapped in the question, answer, and article's title to create a test set composed of entity names of African-origin.

%% file: sections/03-method.tex
\section{Methods}

\subsection {Data} 
We leverage existing extractive MRC dataset, SQuAD2.0~\citep{rajpurkar-etal-2018-know}, where we apply our perturbation method on the dev set to create a perturbed dev set that we name AfriSQuAD2. We choose to conduct our evaluation on SQuAD2.0 because 1) SQuAD2.0 is an extractive QA dataset, i.e the answers are short and are spans from the passage.
2) Questions or answers are composed of named-entities. This allows us to test the model's capability to answer a question on a novel entity or extract a novel entity as an answer.

\subsection{How representative are the entity names in MRC datasets?}

\begin{figure}[!ht]
    \centering
   \subfloat[]{\includegraphics[width=0.4\columnwidth]{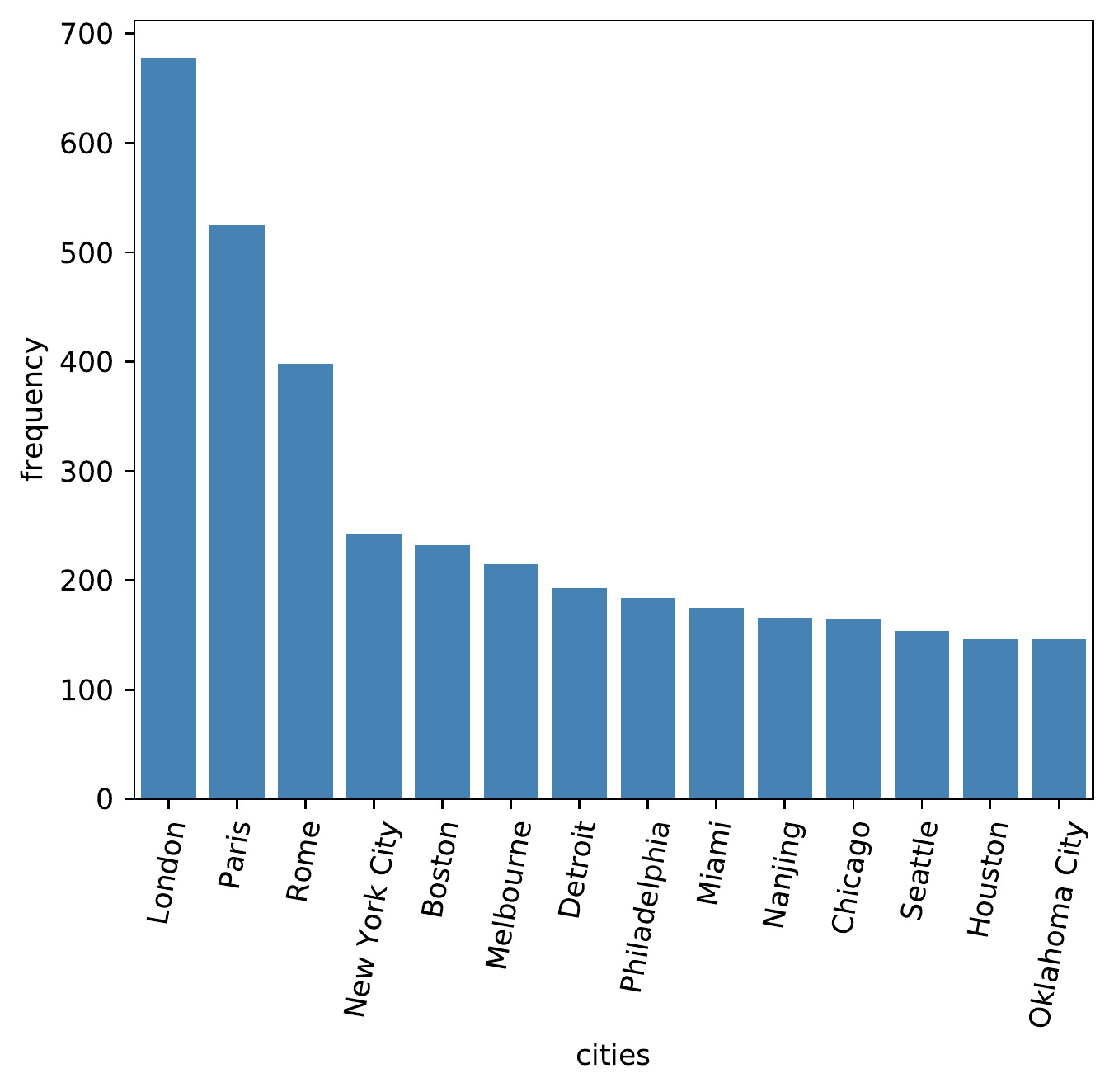}\label{fig:top14:traindist}}
   \subfloat[]{\includegraphics[width=0.4\columnwidth]{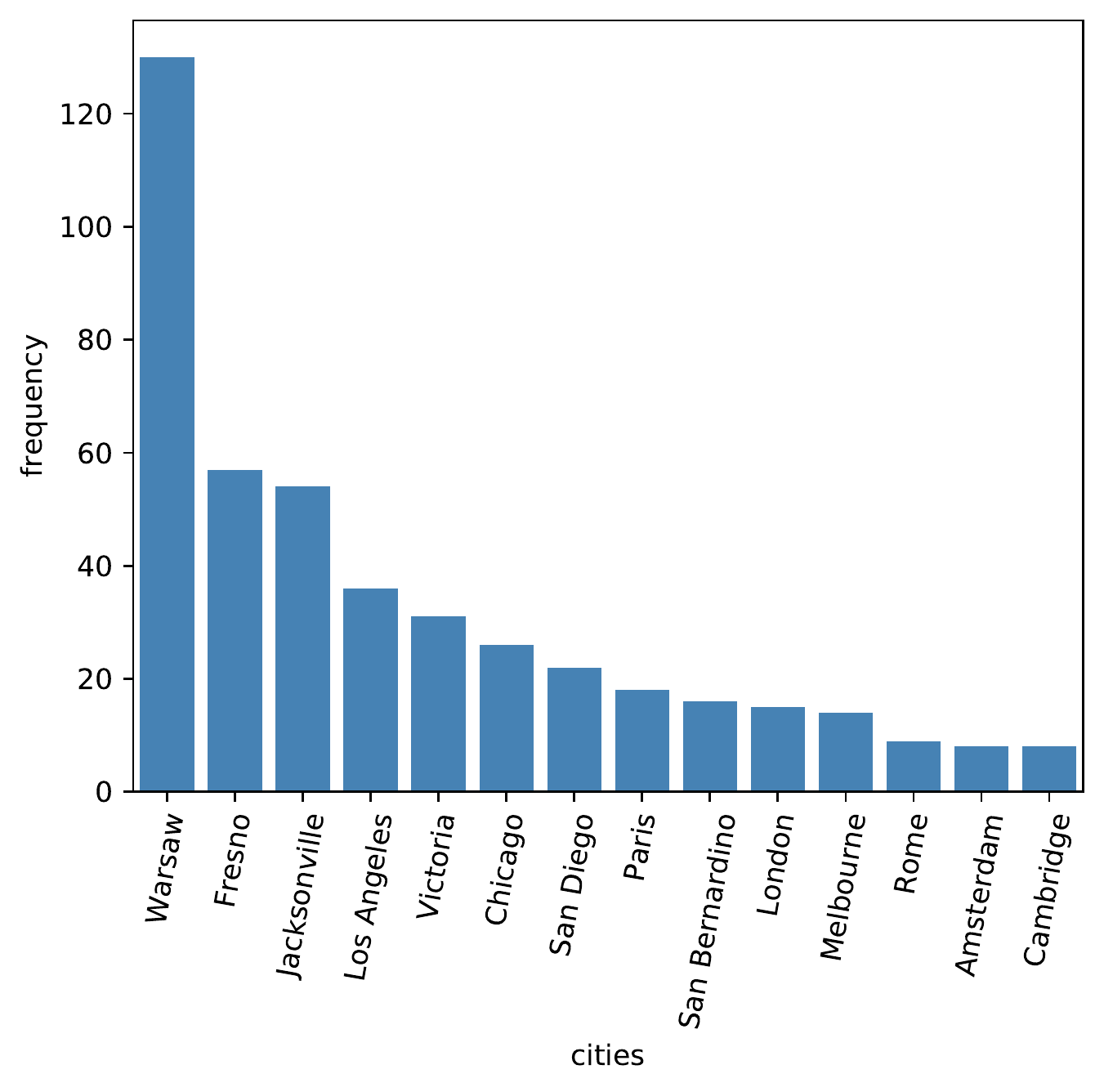}\label{fig:top14:devdist}}\\
   \subfloat[]{\includegraphics[width=0.4\columnwidth]{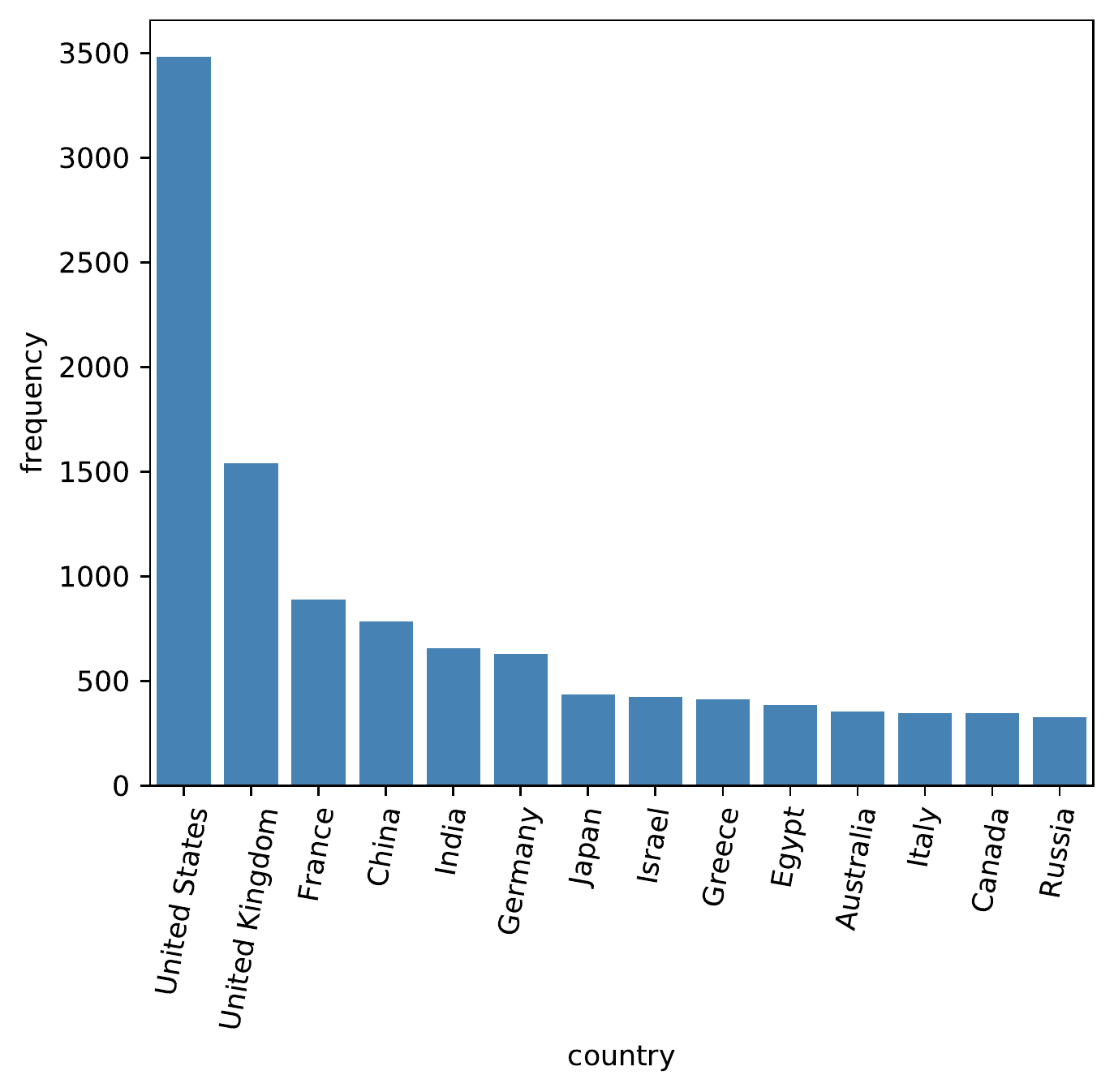}\label{fig:top14:traincountry}}
   \subfloat[]{\includegraphics[width=0.4\columnwidth]{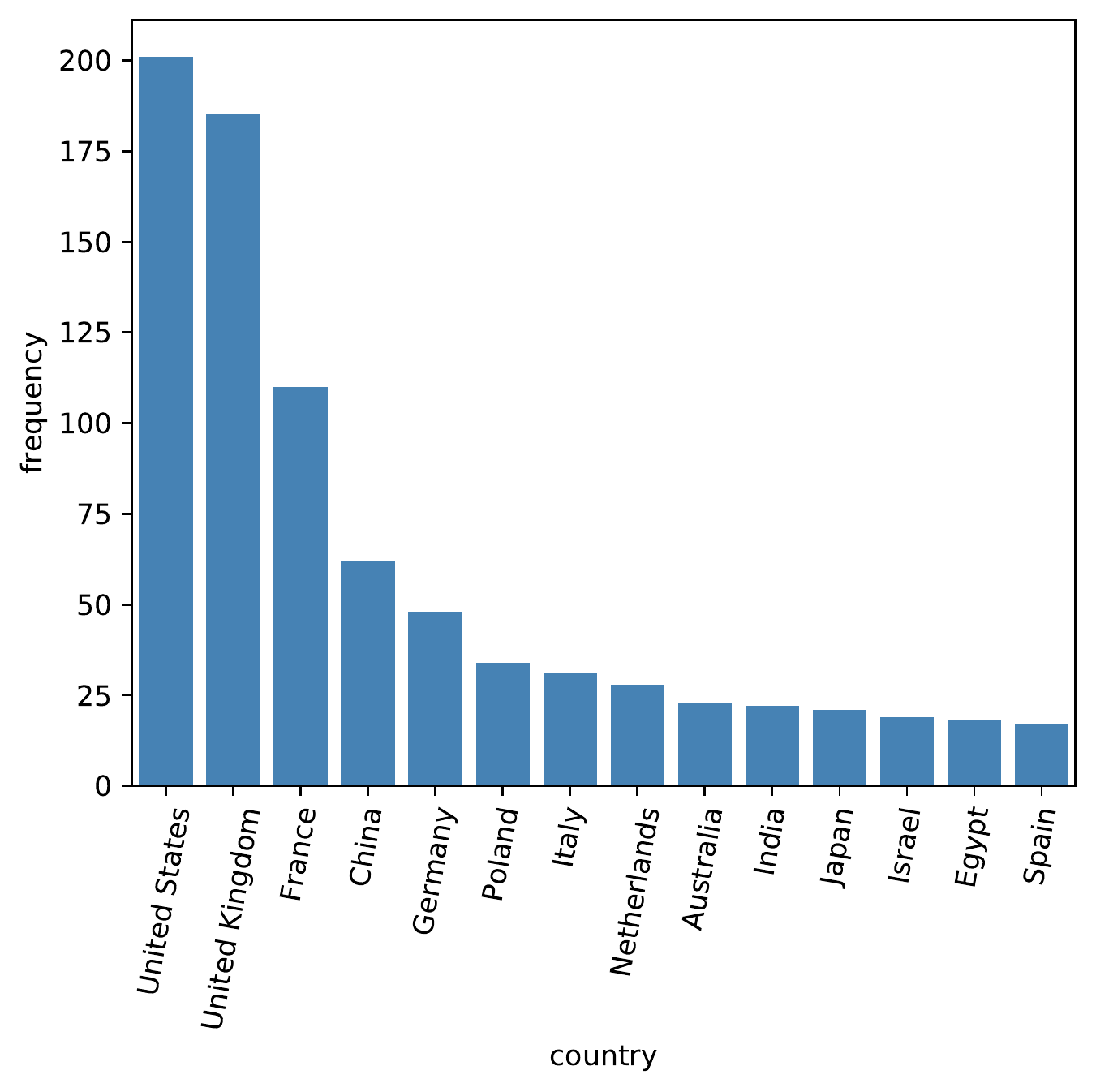}\label{fig:top14:devcountry}}
    \caption{Distribution of top 14 named-entities occurring for city category in (a) train set and (b) dev set, and for country category in (c) train set and (d) dev set. }
    \label{fig:top14}
\end{figure}

 To understand how representative are the entity names in MRC datasets such as SQuAD, we analyze the entity types \emph{city} and \emph{country} in SQuAD2.0. We select the most frequent 14 entities in the train and dev set and map them to their relations in Wikidata, especially their geo-political representation. 
Figure~\ref{fig:top14} shows the top 14 entities in the train and dev set for entity type city in Figures~\ref{fig:top14:traindist} and \ref{fig:top14:devdist} respectively, and entity type country in Figures~\ref{fig:top14:traincountry} and \ref{fig:top14:devcountry}. 

We note that 90\% of the entity names are from either Europe or American continent, thus showing how most articles in the SQuAD2.0 and other MRC datasets in general are not representative of low resource regions. In order to create datasets, researchers mostly rely on Wikipedia as a data source.  The data collected is also influenced by the number of articles available in a particular region. In addition, low-resource regions often have a lower proportion of digital resources compared to high-resource regions. This implies that there are less text data available for training MRC models for such low-resource regions. This can result in less robust models and datasets that do not reflect how diverse the world is. In general we note that there is a skewed representation of named entities towards high resource regions compared to Africa in the SQuAD dataset. Motivated by this, we thus propose to study the robustness of MRC models to entity renaming at test time. Specifically we focus on entity names with African origin. In Section~\ref{entityswap} we describe the creation of the perturbed dev set (AfriSQuAD2) and the experiments conducted.

\subsection{MRC Models}
We experiment with three pretrained language models, which have shown comparative performance on the popular SQuAD2.0 benchmark with base and large variations of the models. \textbf{BERT}~\citep{devlin-etal-2019-bert}, a pretrained deep directional encoder trained on English Wikipedia and BookCorpus, with masked language modeling (MLM) and next sentence prediction (NSP) as the pretraining objective. Unlike BERT, \textbf{RoBERTa}~\citep{DBLP:journals/corr/abs-1907-11692} has shown better performance using only the MLM training objective and pretraining on a large diverse corpus. Compared to BERT and RoBERTa, \textbf{DeBERTa}~\citep{DBLP:conf/iclr/HeLGC21} uses a much larger model size and training corpus, which allows it to capture more complex relationships between words and sentences in a language. 

For model evaluation, we use the models fine-tuned on the original SQuAD2.0 training data from Deepset~\footnote{https://www.deepset.ai/}, publicly available on Huggingface~\citep{wolf2020huggingfaces}. For BERT and RoBERTa, we use the uncased and distilled versions respectively.

%% file: sections/04-entswap.tex
\section{Swapping Entity Names}
\label{entityswap}
In this section, we describe \emph{EntSwap},  our method for perturbing an MRC dev set, by renaming entities with named-entities of African origin. We also describe how we generate collections of entity names for substituting the six categories, with the aim to study the model's robustness.

\subsection{Perturbation Method}
\label{NER}
\begin{figure}
    \centering
    \includegraphics[width=0.8\columnwidth]{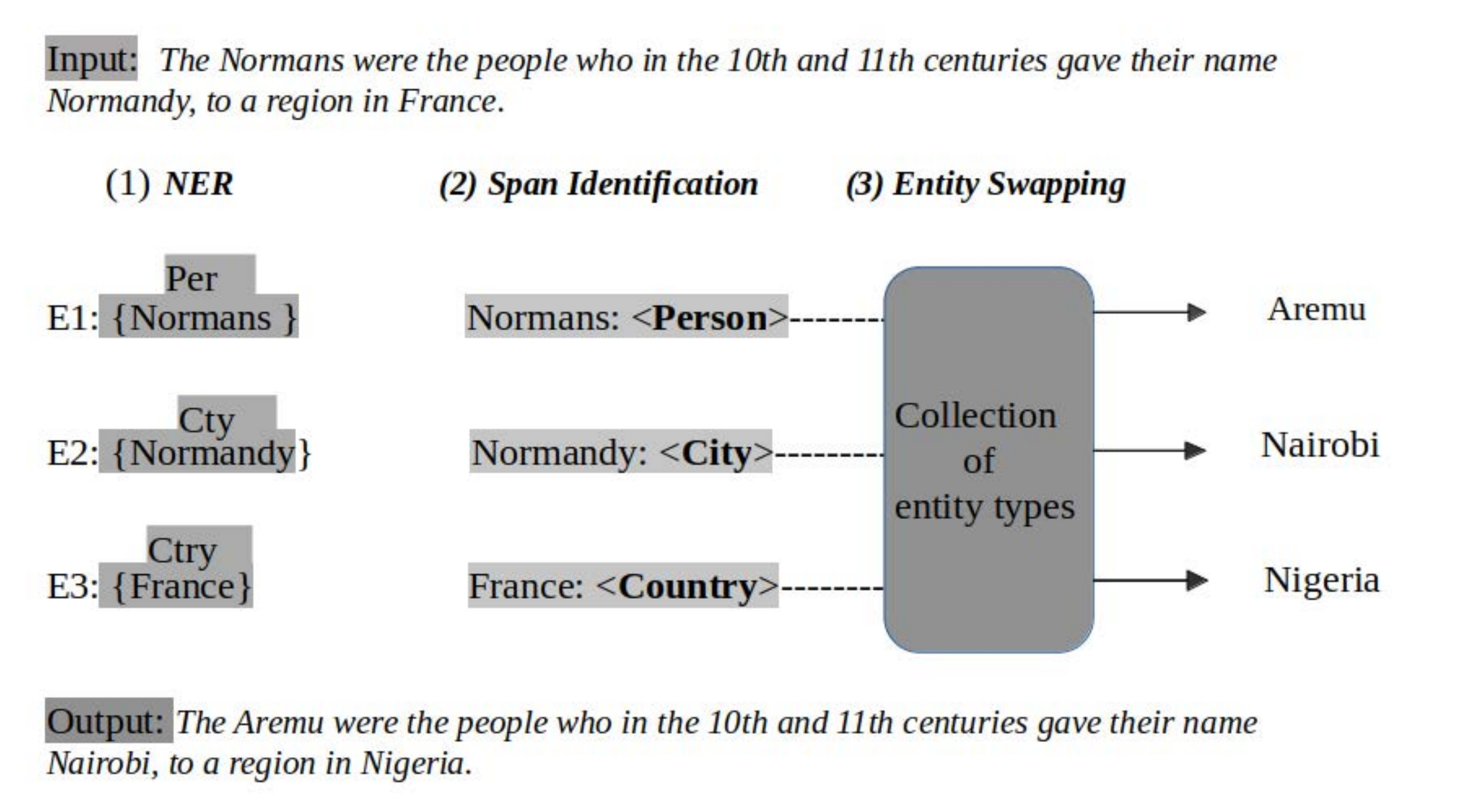}
    \caption{The perturbation method for swapping entity names in an MRC dataset. Per, Cty and Ctry represent Person, City and Country respectively.}
    \label{fig:entswap}
\end{figure}

Previous works create adversarial examples using different techniques such as perturbing by text paraphrasing \citep{iyyer-etal-2018-adversarial}, character-level typos insertion \citep{DBLP:conf/iclr/BelinkovB18}, appending distractors to the input  \cite{jia-liang-2017-adversarial}, and replacing occurrence of certain words in the text with corresponding words \citep{alzantot-etal-2018-generating}. In our work, we create adversarial examples for SQuAD2.0 by randomly swapping specific entities. We use the EntSwap algorithm to create an altered SQuAD2.0 dev set, with a large percentage of entities with Africa-origin.
Figure~\ref{fig:entswap}, shows the perturbation steps.

\paragraph{Step1: Named Entity Recognition.}
In order to identify entities to swap, we run a named-entity recognizer with the Stanza version of the Stanford CoreNLP tool~\citep{DBLP:conf/acl/QiZZBM20} on the context, questions, answers, and titles. Named-Entity Recognition (NER) has proven to be a core component in question-answering tasks, especially extractive question-answering. We use the Stanford CoreNLP tool because of its ability to identify individual GPE entities (i.e., country and city) unlike other tools which have entities country and city categorized as GPE tag.
 In this work, we identify six entity types: \textit{Person }, \textit{Country }, \textit{City }, \textit{Location }, \textit{Organization } and \textit{Nationality }. We mainly focus on these entity types because of their frequent appearances in the question or answer and high possibility of containing valid names.

  \begin{table}[!t]
    \centering
    \begin{tabular}{l c c}
    \toprule
       Category   & Train & Dev  \\
       \midrule
        person &  50706 & 2563 \\
        organization & 27550 & 2041\\
        location & 22327 & 1581\\
        city & 15529 & 1114 \\
        country & 20633 & 1184 \\
        nationality & 16792 & 1104 \\
    \bottomrule
    \end{tabular}
    \caption{Distribution of named-entities by categories in the train and dev sets of the experimental dataset}
    \label{tab:no.entity}
\end{table}

 \paragraph{Step2: Span Identification.}
 The span of an answer in SQuAD2.0 dataset has a start and end position. To identify the span of an entity name to be swapped, we assign the start and end positions of the identified entity. With these positions, we ensure the entity is swapped with another entity of equal length. The number of perturbable spans is shown in Table~\ref{tab:no.entity}, for the train and dev sets. We only swap entities in the dev set.

 \paragraph{Step3: Entity sampling and swapping.}
After the NER tool has identified the entity span to be swapped, we obtain the entity name to be swapped from a collection of entities of the same entity type. The entity name is randomly sampled from a set of entity names from the collection of the same entity type. 

Given a candidate entity name for each pertubable span, we do a string match on the context, question, answer, and title. If an entity occurs more than once in the same context, it is replaced by the same entity name in all instances. An entity name appearing in more than one context may or may not be replaced with the same entity name. To cater for entities that are inflections of another entity, we string-match the main entity and maintain the inflection. For example, \emph{normans} is an inflection of \emph{norman}, so we substitute \emph{norman} with \emph{aremu} and \emph{normans} with \emph{aremus}.

For \emph{person}, to ensure most of the selected names are of African origin, we select the second name or second and third names.  This is because we aim to evaluate our chosen models on novel (African) entities. Most first names are English names, making them not suitable for selection. 

\subsection{Collection of entity names} 
\label{GG}

Using the pre-defined categories in subsection \ref{NER}, we curated named entities from the Wikidata knowledge graph~\citep{DBLP:journals/cacm/VrandecicK14} of African origin.
 We create the collections by extracting canonical names (e.g. Kenya, Abidemi, Tripoli) of six different named-entities. We use SPARQL queries to search Wikidata for named-entities of each category type using relations such as:
 
        \emph{`COUNTRIES IN AFRICA'}, \emph{`CITY OF A COUNTRY, A COUNTRY IN AFRICA'}, and 
        
        \emph{`PERSON BORN IN A CITY, A CITY IN A COUNTRY, A COUNTRY IN AFRICA'}. 
        
To ensure that there are no duplicate entries, we removed all entities with the same Qid occurring more than once. All entities represented with Qids inplace of an entity-name are also manually deleted. We collected a total of six categories and save each entity type into a separate csv file. 
\subsection{Entity Swapping Quality}
The performance of the MRC models is dependent on the quality of our perturbation method. We therefore randomly sample 50 contexts from the dev set and manually check for the quality of the perturbed spans. We evaluate the accuracy of step 2 and 3, that is identifying the perturbable span and swapping the span with the novel entity for categories \textit{Person, Country} and \textit{City}. Results reported in Table~\ref{tab:entswapquality}, show that our method gets acceptable accuracy, thus confirming the quality of the perturbed example.

\begin{table}[h]
    \centering
    \begin{tabular}{l ccc}

\toprule
        & \multicolumn{3}{c}{Accuracy} 
     \\
     \cmidrule(r){2-4}

        Step  & PERSON & COUNTRY & CITY   \\
         \midrule
       Span identification   & 94.32 & 92.17 & 91.26\\
       \EntSwap &88.85  & 96.37 & 94.61 \\
       \bottomrule
    \end{tabular}
    \caption{Accuracy of the two key steps in our perturbation method on 50 randomly sampled contexts from AfriSQuAD2. We report the percentage accuracy.}
    \label{tab:entswapquality}
\end{table}

%% file: sections/05-results.tex
\section{Results and Analysis}
In this section we report the results of our experiments and provide an in-depth analysis to understand which entity types pose a challenge to robustness of the MRC models. We report the automatic metrics of our evaluation i.e. F1-score and exact match~(EM).

\subsection{How robust are MRC models to entity renaming?} 
We conduct evaluation on the SQuAD2.0 and AfriSQUAD2 dev sets.

\begin{table*}[!ht]
    \centering
    
    \begin{tabular}{l cc cc cc}
    \toprule
         & \multicolumn{2}{c}{DeBERTa-large} & \multicolumn{2}{c}{RoBERTa-large} & \multicolumn{2}{c}{BERT-large}
     \\
     \cmidrule(r){2-3}
     \cmidrule(r){4-5}
      \cmidrule(r){6-7}
          
     Dataset & EM & F1 & EM & F1 & EM & F1
     \\
     \midrule
  SQuAD2.0 & 88.07 & 91.14 & 85.08 & 88.26 & 80.83 & 83.83
   \\
   AfriSQuAD2 & 84.14 & 87.54 & 81.60 & 84.96 & 79.29 & 82.52 \\
   $\triangle$ & \textbf{3.93} & \textbf{3.60} & 3.48 & 3.30 & 1.54 & 1.31 \\
    \bottomrule
    \end{tabular}
    \caption{Performance comparison of several MRC models on the SQuAD2.0 and AfriSQuAD2 dev set. We report the F1-score and exact match for both datasets. We represent the difference between the performance of the models on the two datasets with $\triangle$. We boldface models with the greatest performance drop.}
    \label{tab:modelsperformance}
\end{table*}

\begin{table*}[!t]
    \centering
    
    \begin{tabular}{l cc cc cc}
    \toprule
        Dataset  & DeBERTa& RoBERTa & BERT
     
     \\
     \midrule
  SQuAD2.0 & 88.07/83.84  & 85.08/80.35 & 80.83/75.57
   \\
   AfriSQuAD2 & 84.14/80.32  & 81.60/78.05  & 79.29/74.42  \\
    \bottomrule
    
    \end{tabular}
    \caption{Performance comparison of different MRC model sizes. EM scores for the \emph{LARGE/BASE} variants of models on SQuAD2.0 and AfriSQuAD2 .}
    \label{tab:large/base}
\end{table*}

We report the performance of several models on AfriSQuAD2 and SQuAD2.0 dev set in Tables~\ref{tab:modelsperformance} and \ref{tab:large/base}. From the results, we note that: 1) All models show performance drop on AfriSQuAD2 as compared to the original dev set of SQuAD2.0.
2) BERT-base is the most vulnerable model to AfriSQuAD2 with respect to both metrics. This indicates that models that have high performance on the original dev set also tend to perform better on adversarial examples. 3) Large models suffer less from adversarial attacks compare to base models. This is due to their increased size and capacity. Thus, they have the ability to capture more complex patterns and relationships in a data,  such as identifying an answer span with a novel entity name, which in turn improves their accuracy.

\subsection{Which entity types pose a challenge to the MRC Models?}
\begin{table}[!t]
    \centering

    \begin{tabular}{l c c  }
    \toprule

     Dataset & EM & F1 
     \\
     \midrule
     SQuAD2.0 & 80.83 & 83.83 \\
    AfriSQuAD2  & 79.29 & 82.52 \\
    
    \hline \hline
  AfriSQuAD2 w/o city & 78.54 & 81.90 \\
  AfriSQuAD2 w/o country & 79.57 & 82.92 \\
  AfriSQuAD2 w/o location & 80.17& 83.39  \\
  AfriSQuAD2 w/o nationality & 79.68 & 82.99  \\
  AfriSQuAD2 w/o organization & 80.61 & 83.68  \\
  AfriSQuAD2 w/o person & 80.08 & 83.29   \\
    \bottomrule
    \end{tabular}
        \caption{Comparison of BERT-large performance on different entity types in the AfriSQuAD2 dev set.}
    \label{tab:bertablation}
\end{table}

Although BERT models do not perform well in the MRC task compared to their counterparts, they show a minimal performance drop when evaluated on AfriSQuAD2. 
In Table~\ref{tab:bertablation}, we present the performance of the BERT-large model on a different combination of entity types. AfriSQuAD2-w/o-L, where L represents the entity type implies that we replaced all the other five entity types except entity type L. For example, AfriSQuAD2-w/o-city means we swapped all other entity types except \emph{city}. Swapping entity types Per, Org and Loc prove to be challenging to the MRC models' robustness performance. We note that when these entity types are not present in the AfriSQuAD2 dev set the model performs as close as the AfriSQuAD2 dataset. This indicates that renaming these entity types poses a challenge to the robustness of MRC models. This is likely because for \emph{person names}, we select an entity name from its second entity and avoid the first entity since in most cases it is an English name. This way we ensure most of the names are of African-origin and the models may not have been exposed to during training. This also  applies to Organization and Location entity types, most of the organizations in the collection are small,  local organizations within individual countries, which the model did not have access to during training. In the train and dev set of SQuAD2.0, we note that these entity types are the most frequent in the dataset with 50k Person names, 27k Organization names, and 22k Location names. Thus many of these entities are swapped in the dataset compared to CITY, COUNTRY, and NATIONALITY entity types.

\subsection{Analysis}
We do an analysis based on the BERT model to understand the model’s performance towards AfriSQuAD2 examples. We focus on examples where the models originally predicted the correct answer span but failed on the altered examples. We also focus on cases where the model had a low confidence score compared to the original dataset, even when the model extracts the correct answer span.

\paragraph{Error analysis.}
Table\ref{tab:hasans} shows the EM score of BERT-large on HasAns and NoAns questions. Compared to SQuAD2.0, the performance of BERT on AfriSQuAD2 for HasAns is low in comparison to NoAns~(15.56\% drop). This shows that the model is able to predict with high accuracy when a question is unanswerable, unlike a question with answers.  Hence, we seek to understand why the performance of BERT drops on HasAns questions on AfriSQuAD2. We randomly sample 100 questions and classify them into HasAns~(56\%) and NoAns~(44\%) based on the ground truth. We note that 40\% of the HasAns questions were wrongly predicted as NoAns questions. This is mostly the case when either the question or answer was about a novel entity. In particular for person entity type, an entity-name fully of African-origin accounts for most of the drop in the model's performance.

\begin{table}[!t]
    \centering
    \begin{tabular}{l c c}
    \toprule
        Dataset &  HasAns & NoAns\\
        \midrule
        SQuAD2 & 79.41 & 82.25 \\
        AfriSQuAD2 & 71.50 & 87.06 \\
        \bottomrule
    \end{tabular}
    \caption{EM score of BERT-large on questions with answers~(HasAns) and unaswerable questions~(NoAns).}
    \label{tab:hasans}
\end{table}

\paragraph{ Model success and failure. }
Language models such as BERT are pre-trained on large unstructured text corpora with diverse named-entities. They leverage being exposed to a diverse set of named entities. This is why the performance BERT on AfriSQuAD2 does not drop drastically.  We cannot guarantee the same performance for non-transformer based models, especially those not trained on large corpora with diverse named-entities.

Although we report over 80\% accuracy in detecting the span and swapping an entity in the dev set, we can not quantify the role of data quality in the model performance, and thus some of the performance drops may be attributed to data quality.

%% file: sections/06-conclusion.tex
\section{Conclusion}
In this work, we study the robustness of MRC models when entities are swapped to create test time adversarial examples. In particular, we propose \emph{EntSwap}, a method for swapping entity names from the original dataset along with collections of entities of six different categories of African origin. With this method, we create AfriSQuAD2, by renaming entity names in the original SQuAD2.0 dev set with ones from our collections. We experiment with three popular MRC models using SQuAD2.0 dev set and AfriSQuAD2. Although the original SQuAD format was maintained, we find that the AfriSQuAD2 examples challenge the capability of MRC models to extract the correct answer span when the question is about a novel entity. Swapping of entity types: Person, Organization, and location poses the greatest challenge to MRC models. The drop in performance of the models can be attributed to the over reliance of MRC models on real-world entity knowledge. In addition, we observe that most entity names are from high-resource regions,thus, these models may have not been exposed to a subsample of entities in our collections.

For future work, we would like to extend this study to more datasets created from different sources. For example, we would like to see how these models will perform on a perturbed test set of a dataset created via distant supervision compared to a human-created dataset like SQuAD2.0.